\documentclass[10pt,twocolumn,letterpaper]{article}

\usepackage{colortbl}
\usepackage{pifont}
\usepackage[noend]{algpseudocode}  
\usepackage{booktabs}
\usepackage{multirow}
\usepackage{amsmath,amssymb,amsfonts}
\usepackage[ruled,vlined,linesnumbered,boxed,commentsnumbered]{algorithm2e}

\usepackage[pagenumbers]{cvpr} 

\usepackage[dvipsnames]{xcolor}

\newcommand{\vct}[1]{\boldsymbol{#1}} 
\newcommand{\mat}[1]{\boldsymbol{#1}} 
\newlength\savewidth\newcommand\shline{\noalign{\global\savewidth\arrayrulewidth
  \global\arrayrulewidth 1.5pt}\hline\noalign{\global\arrayrulewidth\savewidth}}

\newcommand{\methodname}{FedOFA\xspace}
\newcommand{\methodwoORname}{FedOFA$^*$\xspace}
\newcommand{\TFAname}{TFA\xspace}
\newcommand{\Intrname}{IntraFA\xspace}
\newcommand{\Intename}{InterFA\xspace}
\newcommand{\Oname}{OR\xspace}
\newcommand{\AGPname}{AGPS\xspace}

\definecolor{cvprblue}{rgb}{0.21,0.49,0.74}
\definecolor{lightgray}{rgb}{0.93,0.93,0.93}
\usepackage[pagebackref,breaklinks,colorlinks,citecolor=cvprblue]{hyperref}

\def\paperID{5326} 
\def\confName{CVPR}
\def\confYear{2024}

\title{Energizing Federated Learning via Filter-Aware Attention}

\author{%
  Ziyuan~Yang$^{1}$,
  Zerui~Shao$^{2}$,
  Huijie~Huangfu$^{1}$,
  Hui~Yu$^{1}$,
  Andrew~Beng~Jin~Teoh$^{3}$
  \\
  Xiaoxiao~Li$^{4}$,
  Hongming~Shan$^{5}$,
  and~Yi~Zhang$^{2,\ast}$\\
  \textsuperscript{1}College of Computer Science, Sichuan University\\
  \textsuperscript{2}School of Cyber Science and Engineering, Sichuan University\\
  \textsuperscript{3}School of Electrical and Electronic Engineering, College of Engineering, Yonsei University\\
  \textsuperscript{4}Department of Electrical and Computer Engineering, The University of British Columbia\\
  \textsuperscript{5}Institute of Science and Technology for Brain-Inspired Intelligence, Fudan University\\
  \texttt{\small cziyuanyang@gmail.com}, \texttt{\small zeruishao@outlook.com}, \texttt{\small huangfu@stu.scu.edu.cn},  \\
  \texttt{\small smileudora@163.com}, \texttt{\small bjteoh@yonsei.ac.kr}, \texttt{\small xiaoxiao.li@ece.ubc.ca}, \\
  \texttt{\small hmshan@fudan.edu.cn}, \texttt{\small yzhang@scu.edu.cn}
}

\begin{document}
\maketitle

\def\thefootnote{$\ast$}\footnotetext{Corresponding author}

\begin{abstract}


Federated learning (FL) is a promising distributed paradigm, eliminating the need for data sharing but facing challenges from data heterogeneity. Personalized parameter generation through a hypernetwork proves effective, yet existing methods fail to personalize local model structures. This leads to redundant parameters struggling to adapt to diverse data distributions. To address these limitations, we propose \methodname, utilizing personalized orthogonal filter attention for parameter recalibration. The core is the Two-stream Filter-aware Attention (\TFAname) module, meticulously designed to extract personalized filter-aware attention maps, incorporating Intra-Filter Attention (\Intrname) and Inter-Filter Attention (\Intename) streams. These streams enhance representation capability and explore optimal implicit structures for local models. Orthogonal regularization minimizes redundancy by averting inter-correlation between filters. Furthermore, we introduce an Attention-Guided Pruning Strategy (AGPS) for communication efficiency. AGPS selectively retains crucial neurons while masking redundant ones, reducing communication costs without performance sacrifice. Importantly, \methodname operates on the server side, incurring no additional computational cost on the client, making it advantageous in communication-constrained scenarios. Extensive experiments validate superior performance over state-of-the-art approaches, with code availability upon paper acceptance.

\end{abstract}

\vspace{-15pt}
\section{Introduction}
\label{sec:intro}

Federated learning (FL) has  arisen as a promising distributed learning paradigm that is strategically crafted to mitigate privacy concerns by facilitating collaborative model training among clients without direct data exchange. However, a primary challenge within the FL framework pertains to the issue of data heterogeneity. The presence of disparate data sources with varying characteristics can notably undermine model performance~\cite{liu2021feddg}.

In conventional FL training, clients collectively strive to train a global shared model~\cite{mcmahan2017communication}. However, the presence of notable data heterogeneity poses a challenge, potentially leading to model degradation upon aggregation~\cite{li2021model}. An effective strategy for this issue involves generating personalized parameters for individual client models through a hypernetwork~\cite{shamsian2021personalized}. Despite its efficacy, these methods necessitate a fixed model structure shared among all clients, impeding the personalization of local model structures to accommodate distinct local data characteristics, particularly in heterogeneous client data~\cite{fallah2020personalized}. Moreover, the generated parameters may exhibit redundancy, lacking the necessary constraints for effective adaptation to specific data distributions. The absence of constraints significantly limits the performance of personalization efforts.

To tackle these challenges, we reconceive the self-attention mechanism ~\cite{attentionallyouneed} and advocate redirecting attention to the generated filters on the server side, as opposed to recalibrating local model features on the client side. Our novel FL approach is called Federated Orthogonal Filter Attention (\methodname). Expanding upon the self-attention mechanism, we introduce a Two-Stream Filter-Aware Attention (\TFAname) module, a key component in \methodname. \TFAname posits that boosting personalized performance can be achieved in two filter-aware attention ways: by enhancing the representative capability of individual filters and by exploring relationships between multiple filters to unveil client-specific implicit structures. Concretely, \TFAname comprises two essential components: Intra-Filter Attention (\Intrname) offers a personalized strategy for selecting critical parameters for individual filters, while Inter-Filter Attention (\Intename ) focuses on discovering implicit client-side model structures by establishing connections between different filters. Consequently, \TFAname can concurrently optimize filters, aligning them with the specific data distribution and enjoying the best of both  worlds.

Directly modeling interconnections among all filters in the network proves impractical due to the substantial computational burden associated with the vast parameter count. As a remedy, we propose an approximation approach by investigating layer-wise relationships to alleviate computational overhead. Diverging from self-attention that employs reshaping operations to create patches and multi-heads, we maintain the integrity of individual filters and employ linear projectors to generate diverse multi-heads. 

It is crucial to emphasize that \TFAname diverges markedly from prior self-attention ~\cite{attentionallyouneed}. Integrating sophisticated attention mechanisms into client-side models demands the refinement of local models, thereby inevitably amplifying computational and communication expenses on the client side. In contrast, the presented \TFAname functions directly on filters, obviating the requirement for model fine-tuning and incurring no supplementary costs on the client side.

Given the potential redundancy of generated parameters across filters, we propose incorporating orthogonal regularization (\Oname) within \methodname to mitigate inter-correlation between filters. This integration of \Oname ensures that filters maintain orthogonality, promoting diversity and enhancing representation capability.

In the training phase, \TFAname progressively enforces parameter sparsity by focusing more on pivotal parameters. This inspired the development of an Attention-Guided Pruning Strategy (\AGPname) to economize communication expenses. \AGPname evaluates the importance of neurons within filters, enabling the tailored customization of models by masking unessential neurons. Diverging from \Intename, \AGPname explicitly seeks to explore personalized local architectures, enabling \methodname to achieve communication efficiency without compromising performance.

\noindent\textbf{Contributions.} Our contributions can be summarized as:
\begin{itemize}

 \item We explore the pivotal role of filters in personalized FL and introduce \methodname to underscore their significance. Furthermore, we introduce \TFAname to recalibrate the personalized parameters in a filter-aware way to adapt specific data distribution.

\item The proposed \methodname enriches filter representation capabilities and uncovers implicit client structures without increasing client-side expenses. Furthermore, we furnish a theoretical convergence analysis for our approach.

 \item  We present an attention-guided pruning strategy aimed at mitigating communication overhead by personalizing the customization of local architectures without degrading the performance. 

\end{itemize}

\section{Related Works}
\label{sec:rela}

\subsection{Federated Learning}

Federated learning, a burgeoning distributed collaborative learning paradigm, is gaining traction for its privacy-preserving attributes, permitting clients to train models without data sharing. A seminal approach, FedAvg~\cite{mcmahan2017communication}, necessitates clients to perform local model training, followed by model aggregation on the server. Addressing the constraint of divergent data distributions among clients, several methods have incorporated proximal elements to tackle model drift, exemplified by FedProx~\cite{li2020federated} and pFedMe~\cite{t2020personalized}.

Recent advancements in personalized FL have emerged to tailor global models to specific clients. FedPer~\cite{arivazhagan2019federated} aimed to create specific classifiers, sharing some global layers. FedBN~\cite{li2020fedbn} alleviated heterogeneity concerns through personalized batch normalization (BN) on the client side. LG-FedAvg~\cite{liang2020think} integrated a local feature extractor and global output layers into the client-side model. 
FedKD~\cite{wu2022communication} distilled the knowledge of the global model into local models to enhance personalization performance. Further innovations combined prototype learning with FL~\cite{tan2022fedproto,huang2023rethinking}.

Concurrently, one effective way to achieve personalization is introducing the hypernetwork to generate client-specific models.
Jang~\etal~\cite{jang2022personalized} proposed to use hypernetwork to modulate client-side models for reinforcement learning. Yang~\etal~\cite{yang2022hypernetwork} explored hypernetworks for customizing local CT imaging models with distinct scanning configurations. Transformer-based personalized learning methods~\cite{li2022fedtp} used  hypernetworks to generate client-specific self-attention layers. However, this method mandates client-side models to be based on transformers.

Additionally, pFedLA~\cite{ma2022layer} personalized weights in client-side models during aggregation via hypernetworks. Nonetheless, it lacks asynchronous training, requiring clients to participate in every round. FedROD~\cite{chen2021bridging} utilized hypernetworks as predictors for local models via distillation loss. Shamsian~\etal~\cite{shamsian2021personalized} employed a shared hypernetwork to generate all client-specific model parameters, referred to as pFedHN. Although these methods have demonstrated impressive performance, there remains room for performance improvement  and communication efficiency, particularly by focusing on filters and customizing local model structures.

\subsection{Attention Mechanism}
The attention mechanism emulates human vision, prioritizing salient elements to allocate focus to crucial areas dynamically. Hu~\etal~\cite{hu2018squeeze} introduced channel attention via the squeeze-and-excitation module. Building upon this, Woo~\etal~\cite{woo2018cbam} amalgamated channel and spatial attention, proposing the convolutional block attention module. Hou~\etal~\cite{hou2021coordinate} incorporated position data into channel attention, naming it coordinate attention. With inspiration from the self-attention mechanism's success in natural language processing \cite{vaswani2017attention}, there is a growing interest in integrating self-attention into computer vision. Dosovitskiy~\etal~\cite{dosovitskiy2020image} demonstrated self-attention's excellence in image recognition. Swin-transformer~\cite{liu2021swin} extended self-attention with a shifted window.

While self-attention delivers promising outcomes, it focuses on features. This implies that integrating it into FL necessitates additional efforts, such as fine-tuning client-side models and augmenting client-side computational costs.
\vspace{-5pt}
\section{Methodology}
\label{sec:meth}

\begin{figure}
    \includegraphics[width=1\linewidth]{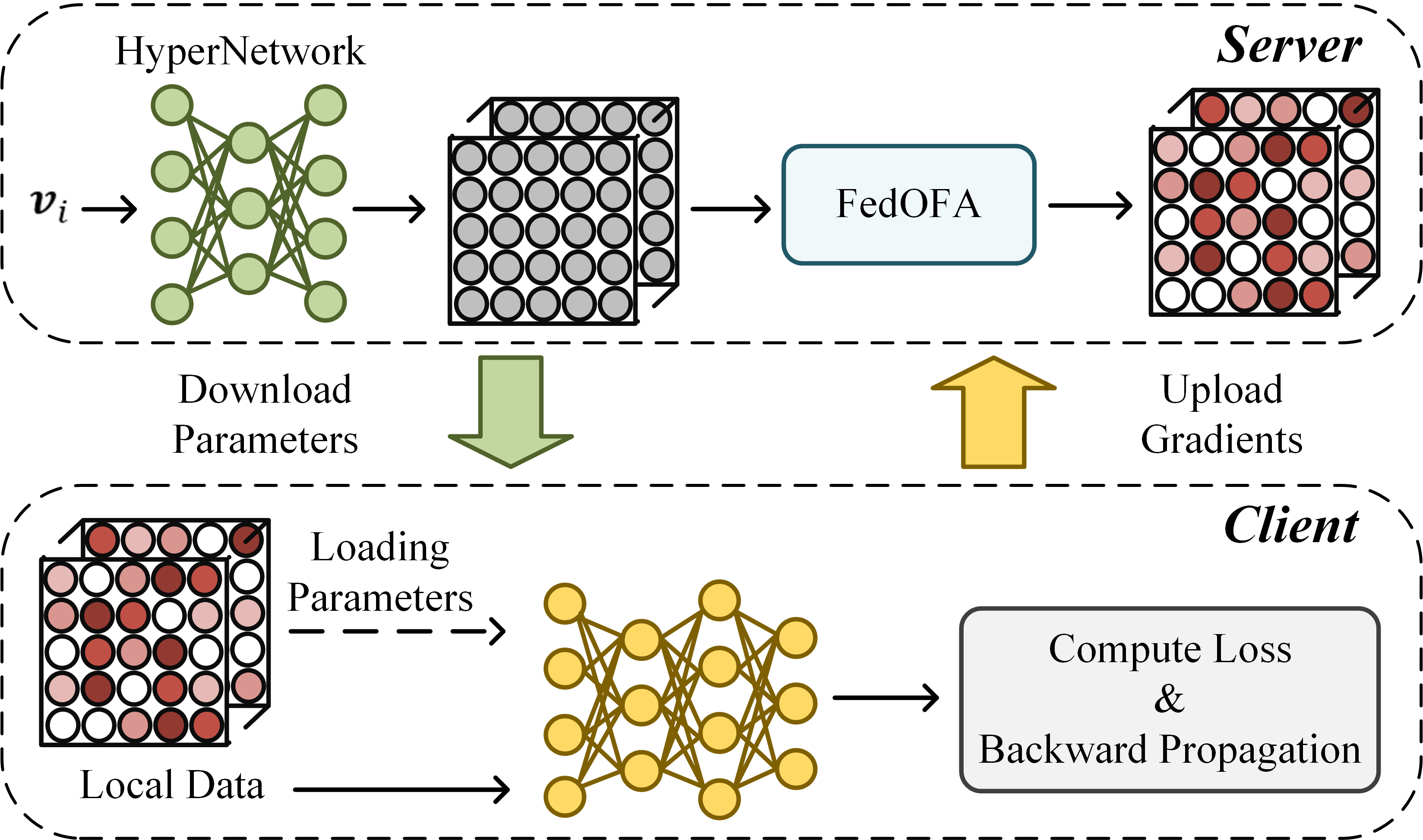}
  \vspace{-10pt}
  \caption{The pipeline of the proposed \methodname. Circles indicate neurons in the filter, with deeper colors indicating higher importance, while white circles are masked neurons.}
  \label{fig:Framework}
  \vspace{-15pt}
\end{figure}

\begin{figure*}
    \includegraphics[width=1\linewidth]{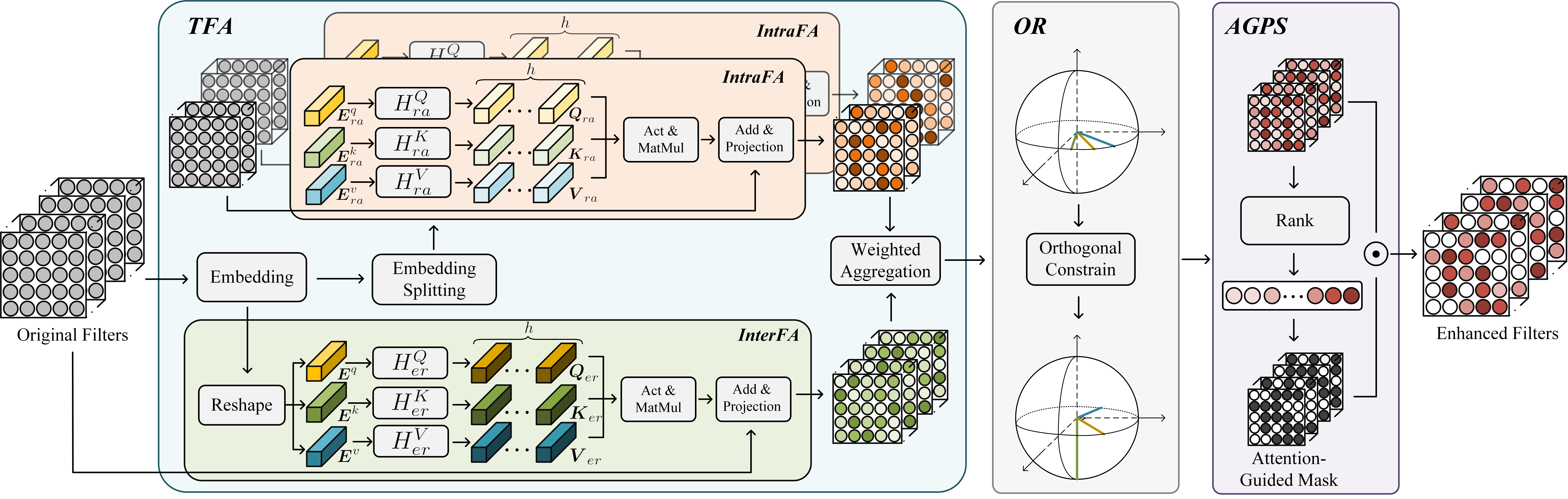}
    \vspace{-15pt}
      \caption{ The proposed \methodname involves a parameter enhancement process through \TFAname, comprised of \Intrname and \Intename. \Intrname is dedicated to augmenting the feature representation capacity of individual filters, while \Intename is geared towards uncovering implicit structures. The \Oname is employed to bolster the diversity of filters. \AGPname is used to mask neurons based on their importance ranking selectively. This process serves the dual purpose of achieving personalized model structures and enhancing communication efficiency.}
  \vspace{-15pt}
  \label{fig:Concept}
\end{figure*}

\subsection{Overview of \methodname}

The proposed \methodname operates on the server and focuses on recalibrating filters to adapt specific data distribution. 
The pipeline of \methodname is depicted in Fig.~\ref{fig:Framework}. Notably, the parameter generation for $i$-th client model is derived from its corresponding embedding vector $\vct{v}_i$ through a hypernetwork. 
All operations within the \methodname are executed on the server and directly manipulate the generated parameters of local models. This design offers two significant advantages for implementation: (\textit{\textbf{i}}) It eliminates the need for fine-tuning hypernetwork architectures and client-side models; and (\textit{\textbf{ii}}) It avoids incurring additional communicational overhead and client-side computational costs, which is particularly vital in resource-constrained scenarios. \methodname comprises three main modules: \TFAname, \Oname, and \AGPname, as illustrated in Fig.~\ref{fig:Concept}. In the following, each of these modules will be detailed in the sequel.

\subsection{Two-Stream Filter-Aware Attention}

In this study, \TFAname is proposed to enhance filter representation capabilities and uncover personalized implicit structures. However, establishing relationships across the entire network incurs prohibitively high computational costs. Consequently, we propose an approach that approximates network-wide relationships layer-by-layer. Suppose there are $n_i$ filters in the $i$-th layer, and $\mat{C}_j^i$ denotes the $j$-th in the $i$-th layer. Meanwhile, we concatenate $\{\mat{C}_1^i, \ldots, \mat{C}_{n_i}^i\}$ to present the parameter of $i$-th layer $\mat{L}^i\in\mathbb{R}^{n_i\times S_{in}\times k\times k}$, where $S_{in}$ and $k$ correspond to the input channel size, and $k$ denotes the kernel size. Similar to existing self-attention techniques, we initially employ learnable embeddings to transform the input into a high-level semantic feature vector denoted as $\mat{E}\in\mathbb{R}^{3\times (n_i \times S_{in}\times k\times k)}$.

\noindent\textbf{Intra-filter attention.}
\Intrname is devised to enhance the representation capabilities of filters, necessitating the division of layer-wise embeddings into filter-aware embeddings. For a typical convolutional filter $\mat{C}\in\mathbb{R}^{S_{in}\times k\times k}$, we perform a split operation on the vector $\vct{E}$ to obtain individual filter embeddings denoted as $\vct{E}_{1},...,\vct{E}_{n_i}$. For conciseness, we represent the filter-aware embedding of a single filter as $\mat{E}_{ra}\in\mathbb{R}^{3\times (S_{in}\times k\times k)}$. Similarly to prior operations, we reshape $\mat{E}_{ra}$ to derive query, value, and key embeddings, denoted as $\mat{E}_{ra}^q, \mat{E}_{ra}^k, \mat{E}_{ra}^v$, for the \Intrname stream.

Prior study has demonstrated the effective performance improvement gained through the multi-head mechanism~\cite{xie2021internal}. However, these methods typically implement multi-head mechanisms by reshaping query, key, and value embeddings. This adaptation is necessitated by the computational overhead associated with establishing direct latent relationships between all features. In this work, we strive to preserve the integrity of each convolution filter while harnessing all parameters to uncover latent relationships within each filter. Fortunately, in comparison to feature dimensions, filter dimensions are generally much smaller, enabling us to treat the entire filter as a patch. Therefore, we introduce a linear projection with random initialization to automatically enhance the diversity by generating multiple heads, as formalized below:
\begin{align}
    \label{align.qifa}\vct{Q}_{ra} &= H_{ra}^Q(\mat{E}_{ra}^q, h),\\
    \label{align.kifa}\vct{K}_{ra} &= H_{ra}^K(\mat{E}_{ra}^k, h),\\
    \label{align.vifa}\vct{V}_{ra} &= H_{ra}^V(\mat{E}_{ra}^v, h),
\end{align}
where $\vct{Q}_{ra}, \vct{K}_{ra}, \vct{V}_{ra}$ denote the query, key, and value vectors of \Intrname, respectively. $\vct{Q}_{ra}, \vct{K}_{ra}, \vct{V}_{ra}\in\mathbb{R}^{h\times (S_{in}\times k\times k)}$. $h$ denotes the number of heads. $H_{ra}^Q(\cdot,h)$, $H_{ra}^K(\cdot,h)$, and $H_{ra}^V(\cdot,h)$ represent the linear projections to generate multi-heads for query, key, and value, respectively. After improving the diversity through the multi-head attention mechanism, the attention map can be acquired from:
\begin{equation}
    \label{equation.map}
    \mat{Out}_{ra} = Att(\mat{Q}_{ra}\mat{K}_{ra})\mat{V}_{ra},
\end{equation}
where $Att(\cdot)$ is the activation function, and the softmax function is used in this paper. A linear projection $P_{ra}(\cdot)$ is used to explore latent relationships among multi-head attention and match the dimension of $C$, which can be formulated as:
\begin{equation}
    \mat{C}_{IntraFA} =  \mat{C} + P_{IntraFA}(\mat{Out}_{ra}),
\end{equation}
where $\mat{C}_{IntraFA}\in\mathbb{R}^{S_{in}\times k\times k}$ is the recalibrated filter.

Accordingly, \Intrname significantly enhances filter representations by capturing the latent connections among all parameters within each filter. Nonetheless, relying solely on \Intrname is insufficient for uncovering the inter-filter relationships necessary to explore the personalized implicit structures within client-side models.

\noindent\textbf{Inter-filter attention.}
Existing hypernetwork-based methods generally require fixed local model structures, but the network architecture significantly affects performance, making a shared structure for all clients impractical. Designing personalized structures for diverse clients based on their local data distributions is cost-prohibitive and architecturally challenging. Some methods attempt to weigh different streams to unveil implicit architectures~\cite{guo2020learning} or explore optimal architectures through exhaustive searches~\cite{lu2022m}, suffering from substantial computational costs.

To address these challenges, we introduce \Intename, a method that customizes client-side structures based on exploring the importance of different filters to the specific data. \Intename approximates network-wide relationships by modeling layer-wise relationships. First, we reshape vector $\vct{E}$ into \Intename's query, value, and key embeddings, i.e., $\vct{E}^q$, $\vct{E}^k$, and $\vct{E}^v$, where $\vct{E}^q, \vct{E}^k, \vct{E}^v\in\mathbb{R}^{1\times(n_i \times S_{in}\times k\times k)}$. 

Like \Intrname, diversity in the embedding vectors is enhanced using $H_{er}^Q(\cdot, h)$, $H_{er}^K(\cdot, h)$, and $H_{er}^V(\cdot, h)$ based on Eqs.~\eqref{align.qifa}, \eqref{align.kifa}, and \eqref{align.vifa}. Then, we derive query, key, and value vectors i.e. $\vct{Q}_{er}$, $\vct{K}_{er}$, and $\vct{V}_{er}$, where $\vct{Q}_{er}, \vct{K}_{er}, \vct{V}_{er}\in\mathbb{R}^{h\times (n_i \times S_{in}\times k\times k)}$. Next, Eq.~\eqref{equation.map} is employed to generate the attention map $\mat{Out}_{er}$ for \Intename, and then we could obtain the final recalibrated layer $\mat{L}_{InterFA}$ as follows:
\begin{equation}
    \label{equation.lefa}
    \mat{L}_{InterFA} =  \mat{L} + P_{InterFA}(\mat{Out}_{er}),
\end{equation}
where $P_{InterFA}(\cdot)$ is a linear projection to learn the implicit relationship among multi-head attentions and to match the dimension.

\Intename operates to establish relationships among distinct filters, thus facilitating the exploration of personalized implicit structures to the specific data. It treats each filter as a cohesive patch, preserving its integrity.

\Intrname and \Intename function independently, concentrating on complementary aspects to enhance model performance. Specifically, the $i$-th layer comprising $n_i$ filters requires $n_i$ \Intrname modules and only one \Intename module.
We introduce \TFAname, where \Intrname and \Intename run as parallel streams to combine both merits. Given the inherent link between enhancing filters and uncovering implicit structures, \TFAname can potentially optimize performance through joint optimization.

As the value embedding $\vct{E}^v$ characterizes input contents and both streams involve filter-aware attention, we can reshape $\vct{V}_{ra}$ from $\vct{V}_{er}$ to reduce computational overhead. The keys and queries operate independently in the two streams for distinct purposes. After enhancing each filter via \Intrname, we concatenate them to obtain the enhanced layer $\mat{L}_{IntraFA}$. Ultimately, the recalibrated layer is derived through weighted aggregation of the two streams as follows:
\begin{equation}
    \label{equation.TFA}
    \mat{L}_{\TFAname} = w\mat{L}_{IntraFA}+ (1-w)\mat{L}_{InterFA},
\end{equation}
where $w$ is the weight of \Intrname, which is empirically set to 0.5 in this study.

\subsection{Orthogonal Regularization}

\TFAname presents an opportunity to recalibrate the network to the specific data by improving filters and exploring implicit structures. However, it lacks consideration for the independence among filters. To mitigate redundancy and promote filter diversity, \Oname is introduced~\cite{xie2017all}. Leveraging orthogonality in linear transformations, OR preserves energy and reduces redundancy in the model's filter responses~\cite{kim2022revisiting}. \Oname employs a penalty function that accounts for orthogonality among filters in each layer with minimal impact on computational complexity.

Specifically, we first reshape the filter responses $\mat{L}_{TFA}$ into a matrix $\mat{O}\in\mathbb{R}^{u\times h}$, where $u$ represents the output size, and $h=S_{in} \times k\times k$. The optimization objective for $\mat{L}$ can be expressed as $\mat{O}^{\top}\mat{O}=\mat{I}$, where $\mat{I}$ denotes the identity matrix. This formulation aims to enforce orthogonality among filters within each layer, ensuring their maximal independence. $\mathcal{O}$ denotes the set of network weights. Then, \Oname can be formally defined as:
\begin{equation}
\label{eq:or}
   \frac{\lambda}{|\mathcal{O}|} \sum_{\mat{O}\in \mathcal{O}}\left\|\mat{O}^{\top}\mat{O}-I_{h}\right\|_{F}^{2},
\end{equation}
where $\lambda$ is the regularization coefficient, which is set to $1e-4$ in this paper. $|\mathcal{O}|$ is the cardinality of $\mathcal{O}$.

To sum up, \Oname effectively enforces orthogonality among filters, enhancing their diversity and reducing filter redundancy. Consequently, this approach can lead to further improvements in network performance.

\subsection{Attention-Guided Pruning Strategy}

While \Intename endeavors to uncover implicit structures tailored to various clients, it maintains fixed client-side model structures. Besides, FL often necessitates frequent data transmission between clients and the server, resulting in substantial consumption of communication resources~\cite{azam2021recycling,diao2022semifl}. To address these challenges, we introduce \AGPname, capitalizing on \TFAname's capacity to zero out unimportant parameters and emphasize critical ones. \AGPname customizes the pruning process for unimportant parameters using our attention map to identify an optimal local architecture and promote efficient communication. Initially, \AGPname assesses the significance of different neurons and subsequently prunes a specified percentage ($p$\%) of them, which can be formulated as:
\vspace{-8pt}
\begin{equation}
    M(cor)=
\begin{cases}
    1,  & \text{if ABS}(\mat{L}_{\TFAname}(cor))>T\\
    0,  & \text{otherwise},
\end{cases}
\end{equation}                  
\noindent where $\text{ABS}(\cdot)$ denotes the absolute value operation, $T$ represents the value which ranks at $p\%$ in the ascending sorted sequence of $\text{ABS}(\mat{L}_{\TFAname})$, and $cor$ is the coordinate index. Finally, we can get the personalized mask $\mat{M}$.

 We mask unimportant parameters, excluding them from the communication process. This operation can be expressed as $\mat{L}_{\TFAname} \odot \mat{M}$, where $\odot$ signifies the element-wise dot product. Consequently, the transmitted neurons are thoughtfully selected to align with the salient aspects of local data, as determined by our filter-aware attention mechanism. This approach effectively represents a neuron-wise pruning strategy for tailoring local structures based on the specific local data, leading to significant reductions in communication costs compared to transmitting all parameters. The pseudo-code of \methodname is provided in the section of Supplementary Material. 
\section{Model Analysis}

In this section, we propose a theoretical analysis of our methods from two distinct perspectives. Firstly, we establish that our proposed filter-aware attention can be approximated by other feature-aware attention mechanisms. Subsequently, we furnish a proof of convergence for our method.

Let $h(\cdot; \phi)$ be the hypernetwork parameterized by $\phi$, $f(\cdot; \theta_i)$ represent the $i$-th client-side network parameterized by $\theta_i$, and $g(\cdot; \varphi)$ indicate the attention module parameterized by $\varphi$. We denote the training set in the $i$-th client, generated by the distribution $P_i$, as $D_{i}=(\mat{X}_{i}, \vct{y}_{i})$, where $\mat{X}_{i}$ and $\vct{y}_{i}$, respectively denote the training samples and their corresponding labels on the $i$-th client.

\noindent\textbf{Relationship to feature-aware attention.} 
For the $i$-th client, the forward process in feature-aware attention can be represented as $g(f(\mat{x}_i; \theta_i); \varphi)$, where $\theta_i=h(\vct{v}_{i};\phi)$. In contrast, our filter-aware attention formulates the forward process as $f(\mat{x}_i; g(\theta_i; \varphi))$. Assuming the hypernetwork and attention modules are linear models, we have $\theta_i=\mat{W}\vct{v}_{i}$, where $\phi:=\mat{W} \in \mathbb{R}^{d \times k}$, $\varphi:= \mat{A} \in \mathbb{R}^{k \times k}$ for filter-aware attention, and $\varphi:= \mat{A} \in \mathbb{R}^{d \times d}$ for feature-aware attention.

Consistent with the assumption in \cite{shamsian2021personalized}, we further assume $\mat{X}_i^{\top}\mat{X}_i=\mat{I}$ in this study, which can be achieved through data whitening techniques~\cite{genkin2022biological}. In the case of feature-aware attention, we define the empirical risk minimization (ERM) as $\bar{\theta}_{i}=\arg \min _{\theta \in \mathbb{R}^{d}}\left\|\mat{A}\mat{X}_{i} \theta-\vct{y}_{i}\right\|^{2}$, with the optimal solution $\bar{\theta}_{i}=
\mat{X}_i^{\top}\mat{A}^{\top}\vct{y}_i$. Due to OR, $\mat{A}$ can be optimized over matrices with orthonormal columns, i.e., $\mat{A}^{\top}\mat{A}=\mat{I}$. Then, the ERM can be expressed as:
\begin{equation}
    \label{equation.feature}
    \underset{\theta_{i}}{\arg \min}\left(\mat{A}\mat{X}_{i} \theta_{i}-\vct{y}_{i}\right)^{\top}\left(\mat{A}\mat{X}_{i} \theta_{i}-\vct{y}_{i}\right).
\end{equation}

Eq.~\eqref{equation.feature} can be expressed as $\arg \min _{\theta_{i}}\left\|\theta_{i}-\bar{\theta}_{i}\right\|_{2}^{2}$. For the proposed filter-aware attention, the ERM is formulated as $\bar{\theta}_{i}=\arg \min _{\theta \in \mathbb{R}^{d}}\left\|\mat{X}_{i}\mat{A} \theta-\vct{y}_{i}\right\|^{2}$. The optimal solution can be expressed as $\bar{\theta}_{i}=\mat{A}^{\top}\mat{X}_i^{\top}\vct{y}_i$. The ERM solution of the proposed filter-aware attention can be formulated as:
\begin{equation}
    \label{equation.filter}
    \underset{\theta_{i}}{\arg \min} \left(\mat{X}_{i}\mat{A} \theta_{i}-\vct{y}_{i}\right)^{\top}\left(\mat{X}_{i}\mat{A} \theta_{i}-\vct{y}_{i}\right). 
\end{equation}

To wrap it up, we derive the optimal solution by expanding Eq.~\eqref{equation.filter} as $\arg \min _{\theta_{i}}\left\|\theta_{i}-\bar{\theta}_{i}\right\|_{2}^{2}$. Notably, the ultimate optimization objectives of both filter-aware and feature-aware attention are equivalent. Hence, the proposed filter-aware attention approximates feature-aware attention. However, feature-aware attention operates on clients, increasing computational and communication costs for parameter uploads and downloads. In contrast, the proposed filter-aware attention mechanism operates on the server, directly enhancing filters and eliminating additional computational and communication costs on the client side. It is worth noting that all feature-aware attention necessitates fine-tuning client-side models, making another advantage of filter-aware methods evident: the absence of a need for fine-tuning client-side models.

\noindent\textbf{Convergence analysis.}  
In this section, we delve into the analysis of the convergence properties of the proposed method. Within our hypernetwork-based FL framework, the parameter $\vartheta_i$ of $f_i(\cdot)$ in the $i$-th client is a function of $\phi$ and $\varphi$. To eliminate ambiguity, we introduce the definitions $\theta = h(\cdot,\phi)$ and $\vartheta = g(h(\cdot,\phi),\varphi)$. This allows us to formulate the parameter generation process for the $i$-th client-side model as $\vartheta_i = g\left(h\left(\boldsymbol{v}_{i}, \phi\right); \varphi\right)$. In our framework, the personalized client-side parameters are generated by the hypernetwork and the proposed filter-aware attention modules. Consequently, the core objective of the training process is to discover the optimal values for $\varphi$ and $\phi$.

Let $\mat{V}$ denote the matrix whose columns are the clients embedding vectors $\vct{v}_i$. We denote the empirical loss of the hypernetwork as $\hat{\mathcal{L}}_{D}(\mat{V}, \phi, \varphi)=\frac{1}{n} \sum_{i=1}^{n} \frac{1}{m} \sum_{j=1}^{m} \ell_{i}\left(\mat{x}_{j}^{(i)}, \vct{y}_{j}^{(i)} ; \vartheta_i\right)$. Based on this empirical loss, we formulate the expected loss as $\mathcal{L}(\mat{V}, \phi, \varphi) = \frac{1}{n} \sum_{i=1}^{n} \mathbb{E}_{P_{i}}\left[\ell_{i}\left(\mat{x},\vct{y};\vartheta_i\right)\right]$.

To initiate our analysis, we first assume that the parameters of the hypernetwork, attention module, and embeddings are bounded within a ball of radius $R$. We establish five Lipschitz conditions~\cite{azam2021recycling}, which can be expressed as:
\vspace{-5pt}
\begin{align}
    &\left\|\ell_{i}\left(\mat{x}, \vct{y}, \vartheta_{1}\right)-\ell_{i}\left(\mat{x}, \vct{y}, \vartheta_{2}\right)\right\| \leq \beta\left\|\vartheta_{1}-\vartheta_{2}\right\|, \\
     &\left\|h(\vct{v}, \phi)-h\left(\vct{v}, \phi^{\prime}\right)\right\| \leq \beta_{h}\left\|\phi-\phi^{\prime}\right\|, \\
    &\left\|h(\boldsymbol{v}, \phi)-h\left(\vct{v}^{\prime}, \phi\right)\right\| \leq \beta_{v}\left\|\vct{v}-\vct{v}^{\prime}\right\|, \\
    &\left\|g(\theta, \varphi)-g\left(\theta, \varphi^{\prime}\right)\right\| \leq \beta_{g}\left\|\varphi-\varphi^{\prime}\right\|,\\
    &\left\|g(\theta, \varphi)-g\left(\theta^{\prime}, \varphi\right)\right\| \leq \beta_{\theta}\left\|\theta-\theta^{\prime}\right\|.
\end{align}

Similar to the scenario analyzed in~\cite{shamsian2021personalized}, consider parameters $\vartheta_i$ and $\vartheta_i^{\prime}$, generated by sets of values $\vct{v}_1, \ldots, \vct{v}_n$, $\phi$, and $\varphi$, and $\vct{v}_1^{\prime}, \ldots, \vct{v}_n^{\prime}$, $\phi^{\prime}$, and $\varphi^{\prime}$, respectively. In this context, the distance $d$ between the output of the loss function for the $i$-th client-side can be expressed as:
\begin{equation}
    \small
    \begin{split}
        & d\left(\left(\vct{v}_{1},\ldots,\vct{v}_{n},\phi,\varphi\right),\left(\vct{v}_{1}^{\prime},\ldots,\vct{v}_{n}^{\prime},\phi^{\prime},\varphi^{\prime}\right)\right)= \\
        & \underset{\mat{x}_{i},\vct{y}_{i} \sim P_{i}}{\mathbb{E}}\left[\frac{1}{n}\left|\sum\ell_{i}\left(\mat{x}_{i},\vct{y}_{i},\vartheta_i\right)-
        \sum\ell_{i}\left(\mat{x}_{i},\vct{y}_{i},\vartheta_i^{\prime}\right)\right|\right],
    \end{split}
\end{equation}
\noindent where $\vartheta_i^{\prime} = g\left(h\left(\boldsymbol{v}_{i}^{\prime}, \phi^{\prime}\right);\varphi^{\prime}\right)$. Based on the triangle inequality and the above Lipshitz assumptions, we can get the inequality about $d$ as follows:
\begin{equation}
    \small
    \begin{split}
        & d\left(\left(\vct{v}_{1},\ldots,\vct{v}_{n},\phi,\varphi\right),\left(\vct{v}_{1}^{\prime}, \ldots, \vct{v}_{n}^{\prime}, \phi^{\prime}, \varphi^{\prime}\right)\right) \\
        & \leq \sum\frac{1}{n} \underset{\mat{x}_{i},\vct{y}_{i} \sim P_{i}}{\mathbb{E}}\left[\left|\ell_{i}\left(\mat{x}_{i},\vct{y}_{i},\vartheta_i\right)-\ell_{i}\left(\mat{x}_{i},\vct{y}_{i},\vartheta_i^{\prime}\right)\right|\right] \\
        & \leq \beta\left\|\vartheta_i-\vartheta_i^{\prime}\right\|
        \leq \beta\left\|\vartheta_i-\Tilde{\vartheta_i^{\prime}}\right\| + \beta\left\|\Tilde{\vartheta_i^{\prime}}-\vartheta_i^{\prime}\right\| \\
        & \leq \beta \cdot \beta_{g}\left\|\varphi-\varphi^{\prime} \right\| + \beta \cdot \beta_{\theta}\left\|\theta_i-\theta_i^{\prime}\right\| \\
        & \leq \beta \cdot \beta_{g}\left\|\varphi-\varphi^{\prime}\right\| + \beta \cdot \beta_{\theta} \cdot \beta_{h} \left\|\phi-\phi^{\prime}\right\| + \\
        & \quad\ \beta \cdot \beta_{\theta} \cdot \beta_{v} \left\|\boldsymbol{v}_{i} -\boldsymbol{v}_{i}^{\prime}\right\|,
    \end{split}
\end{equation}
\noindent where parameter $\theta_i^{\prime}$ is generated by $\boldsymbol{v}_1^{\prime}, \ldots, \boldsymbol{v}_n^{\prime}, \phi^{\prime}$, the parameters $\Tilde{\theta_i^{\prime}}$ and $\Tilde{\vartheta_i^{\prime}}$ are intermediate variables generated by $\boldsymbol{v}_1^{\prime}, \ldots, \boldsymbol{v}_n^{\prime}, \phi$ and $\boldsymbol{v}_1^{\prime}, \ldots, \boldsymbol{v}_n^{\prime}, \phi^{\prime}, \varphi$, respectively. We denote that $\theta_i^{\prime} = h\left(\boldsymbol{v}_{i}^{\prime}, \phi^{\prime}\right)$, $\Tilde{\theta_i^{\prime}} = h\left(\boldsymbol{v}_{i}^{\prime}, \phi\right)$ and $\Tilde{\vartheta_i^{\prime}} = g\left(h\left(\boldsymbol{v}_{i}^{\prime}, \phi^{\prime}\right);\varphi\right)$. 

We note that the filter-aware attention module does not disrupt the convergence and can be regarded as a plug-and-play module.

\vspace{-5pt}
\begin{table*}[]
\centering
\caption{Test accuracy ($\pm$ STD) over 10, 50, and 100 clients on the CIFAR-10 and CIFAR-100. \methodwoORname and \methodname indicates our method without and with \Oname, respectively.}
\vspace{-5pt}
\begin{tabular}{rccccccc}
\shline
\multirow{2}{*}{Method} & \multicolumn{3}{c}{CIFAR-10} &                                                     & \multicolumn{3}{c}{CIFAR-100}                                                     \\ \cline{2-4} \cline{6-8}
          & 10                        & 50                        & 100                       && 10                        & 50                        & 100                       \\ \hline 
      \rowcolor{lightgray}
Local     & 86.46 $\pm$ 4.02              & 68.11 $\pm$ 7.39              & 59.32 $\pm$ 5.59              && 58.98 $\pm$ 1.38              & 19.98 $\pm$ 1.41              & 15.12 $\pm$ 0.58              \\

FedAvg~\cite{mcmahan2017communication}    & 51.42 $\pm$ 2.41              & 47.79 $\pm$ 4.48              & 44.12 $\pm$ 3.10              && 15.96 $\pm$ 0.55              & 15.71 $\pm$ 0.35                & 14.59 $\pm$ 0.40                \\
\rowcolor{lightgray}
FedProx~\cite{li2020federated}   & 51.20 $\pm$ 0.66    & 50.81 $\pm$ 2.94                & 57.38 $\pm$ 1.08                && 18.66 $\pm$ 0.68                         & 19.39 $\pm$ 0.63                & 21.32 $\pm$ 0.71                \\
MOON~\cite{li2021model} & 50.98 $\pm$ 0.73 & 53.03 $\pm$ 0.53  & 51.51 $\pm$ 2.18  && 18.64 $\pm$ 1.02 & 18.89 $\pm$ 0.54  & 17.66 $\pm$ 0.47
\\
\rowcolor{lightgray}
FedNova~\cite{wang2020tackling} & 48.05 $\pm$ 1.32 & 51.45 $\pm$ 1.25 & 47.19 $\pm$ 0.46  && 16.48 $\pm$ 0.86  & 17.91 $\pm$ 0.61  & 17.38 $\pm$ 0.53  \\
LG-FedAvg~\cite{liang2020think} & 89.11 $\pm$ 2.66              & 85.19 $\pm$ 0.58              & 81.49 $\pm$ 1.56              && 53.69 $\pm$ 1.42              & 53.16 $\pm$ 2.18              & 49.99 $\pm$ 3.13              \\
\rowcolor{lightgray}
FedBN~\cite{li2020fedbn}     &  90.66 $\pm$ 0.41                           &  87.45 $\pm$ 0.95                & 86.71 $\pm$ 0.56        &&  50.93 $\pm$ 1.32                          &  50.01 $\pm$ 0.59      &  48.37 $\pm$ 0.56           \\
pFedMe~\cite{liang2020think}    & 87.69 $\pm$ 1.93              & 86.09 $\pm$ 0.32              & 85.23 $\pm$ 0.58              && 51.97 $\pm$ 1.29              & 49.09 $\pm$ 1.10              & 45.57 $\pm$ 1.02   
                       \\
\rowcolor{lightgray}
FedU~\cite{dinh2021fedu}      & -                         & 80.60 $\pm$ 0.30                & 78.10 $\pm$ 0.50                && -                         & 41.10 $\pm$ 0.20                & 36.00 $\pm$ 0.20   
\\
FedPer~\cite{arivazhagan2019federated}    & 87.27 $\pm$ 1.39              & 83.39 $\pm$ 0.47              & 80.99 $\pm$ 0.71              && 55.76 $\pm$ 0.34              & 48.32 $\pm$ 1.46                & 42.08 $\pm$ 0.18               \\
\rowcolor{lightgray}
FedROD~\cite{chen2021bridging} & 90.95 $\pm$ 1.90 & 88.17 $\pm$ 0.53 & 84.42 $\pm$ 0.51 && 64.27 $\pm$ 3.80  & 57.22 $\pm$ 0.96 & 45.57 $\pm$ 0.56
\\
FedKD~\cite{wu2022communication} & 92.04 $\pm$ 0.90  & 88.24 $\pm$ 0.26 &  82.76 $\pm$ 0.66  && 66.61 $\pm$ 0.94 & 50.27 $\pm$ 1.58  & 35.92 $\pm$ 0.57 
\\
\rowcolor{lightgray}
FedProto~\cite{tan2022fedproto} & 89.65 $\pm$ 1.29  & 84.71 $\pm$ 1.09  & 81.94 $\pm$ 0.47 && 61.74 $\pm$ 1.23 & 57.94 $\pm$ 0.10 & 52.18 $\pm$ 0.53
\\
pFedHN~\cite{shamsian2021personalized} & 91.18 $\pm$ 0.91  & 87.67 $\pm$ 0.67 & 87.95 $\pm$ 0.68  && 65.99 $\pm$ 0.23   & 59.46 $\pm$ 0.26 & 53.72 $\pm$ 0.57 
\\
\rowcolor{lightgray}
pFedLA~\cite{ma2022layer} & 90.58 $\pm$ 0.88 & 88.22 $\pm$ 0.65 & 86.44 $\pm$ 0.74  && 62.73 $\pm$ 0.72  & 56.50 $\pm$ 0.73 & 51.45 $\pm$ 0.65  
\\ \hline 
\methodwoORname       & \textit{92.57 $\pm$ 0.58}        & \textit{89.29 $\pm$ 0.44}         & \textit{88.64 $\pm$ 0.26}        && \textit{66.73 $\pm$ 0.33}        & \textit{60.17 $\pm$ 0.68}        & \textit{54.84 $\pm$ 0.36}        \\
\rowcolor{lightgray}
\methodname       & \textbf{93.18 $\pm$ 0.72}        & \textbf{90.75 $\pm$ 0.79}         & \textbf{89.42 $\pm$ 0.99}        && \textbf{68.28 $\pm$ 0.34}        & \textbf{61.08 $\pm$ 0.23}        & \textbf{56.25 $\pm$ 0.45}        \\ \shline
\end{tabular}
\label{tab:acc}
\end{table*}

\section{Experiments}
\subsection{Experimental Setting}

We assess our method's performance using two well-known public datasets, CIFAR-10 and CIFAR-100~\cite{krizhevsky2009learning}. To create a heterogeneous client setup akin to the challenging scenario in \cite{shamsian2021personalized}, we diversify clients based on class composition and their local training data size. Specifically, for CIFAR-10, we randomly assign two classes to each client, while for CIFAR-100, ten classes are allocated. The sample ratio for a chosen class $c$ on the $i$-th client is determined as $a_{i, c} / \sum_{j=1}^n a_{j, c}$, where $a_{i, c}$ follows a uniform distribution in the range of 0.4 to 0.6. Here, $n$ represents the total number of clients. This procedure results in clients with varying quantities of samples and classes while ensuring that both local training and testing data adhere to the same distribution. The experiments in other datasets can be found in the section of Supplementary Material.

We adopt a hypernetwork structure consistent with prior studies to ensure fair performance evaluation, featuring three hidden layers and multiple linear heads for each target-weight tensor. We configure the protocol with 5000 communication rounds and 50 local training iterations. For the client-side network, we employ LeNet~\cite{lecun1998gradient}, comprising two convolutional layers and two fully connected layers. 

The experimental setup involves a cluster with a single NVIDIA RTX 3090 GPU, simulating the server and all clients. Implementation is carried out using the PyTorch framework, employing a mini-batch size of 64 and stochastic gradient descent (SGD) as the optimizer with a learning rate of 0.01.

\subsection{Results}
We compare our method with local training (Local) and several state-of-the-art FL methods. All the methods share the same client-side models to ensure the fairness of the comparison. For FedBN, we follow its original settings and add a batch normalization layer after each convolutional layer in client-side models.

\begin{table*}[]
\centering
\vspace{-5pt}
\small
\caption{Experiments about \AGPname over 100 clients on the CIFAR-10 and CIFAR-100.}
\begin{tabular}{rccccccc}
\shline 
  &   baseline     & upper bound                        & $p=70$                       & $p=80$                       & $p=90$                       & $p=95$                       & $p=99$                       \\ \hline
\rowcolor{lightgray}
CIFAR-10 & 87.95 $\pm$ 0.68      & 88.64 $\pm$ 0.26              & 87.93 $\pm$ 0.25              & 87.81 $\pm$ 0.59              & 86.89 $\pm$ 0.42              & 85.81 $\pm$ 0.80              & 83.61 $\pm$ 0.79              \\ 
CIFAR-100 & 53.72 $\pm$ 0.57   & 54.84 $\pm$ 0.36            & 53.44 $\pm$ 0.79              & 53.69 $\pm$ 0.58              & 52.88 $\pm$ 0.37              & 51.54 $\pm$ 0.60                & 46.67 $\pm$ 0.62                    \\ \shline
\end{tabular}
\vspace{-10pt}
\label{tab:communication}
\end{table*}

Tab.~\ref{tab:acc} reports the average test accuracy and the associated standard deviation (STD) for all algorithms. Notably, FedAvg, FedProx, MOON, and FedNova exhibit subpar performance across most scenarios, primarily attributed to their non-personalized FL approach, rendering them ill-equipped to handle data heterogeneity challenges effectively. Among these methods, pFedHN is the baseline hypernetwork-based FL approach, devoid of any attention module. To ensure a level playing field for comparison, \methodname employs the same hypernetwork for parameter generation. 

The results underscore the substantial performance enhancement achieved by our method, accomplished by personalized filter improvements and implicit structure exploration. Compared to other methods, our approach consistently demonstrates competitive performance. Furthermore, the integration of \Oname further refines the performance of \methodname by minimizing filter redundancy and improving the diversity.

To mitigate the formidable communication costs inherent to FL, we introduce \AGPname, a mechanism for the judicious selection of vital parameters. To ensure a fair assessment, we assess \methodwoORname with \AGPname across 100 clients in CIFAR-10 and CIFAR-100 in Tab.~\ref{tab:communication}. We employ the vanilla hypernetwork-based method devoid of \methodname as the baseline and set $p=0$ as the upper bound.

Our findings reveal that our method continues to deliver competitive performance even when pruning a substantial 70\% and 80\% of parameters. This underscores \AGPname's effectiveness in singling out pivotal parameters, curtailing communication costs. However, it's worth noting that our method exhibits a slight drop in accuracy when reducing the parameter count to just 5\%. Conversely, even with only 1\% of the parameters, our method surpasses the performance of other FL methods, as elucidated in Tab.~\ref{tab:acc}.

These findings underscore the dispensability of numerous parameters within the network. We can concurrently enhance communication efficiency and performance by delving into implicit personalized structures. It's imperative to recognize that these insights pertain to the FL setting, wherein all costs scale linearly with the client count. Consequently, optimizing communication efficiency becomes a pivotal concern, mainly when dealing with many clients during training. The ability to achieve competitive performance with as little as 5\%, or even 1\%, of the parameters underscores the acceptability of slight performance degradation in exchange for remarkable efficiency gains.

\vspace{-5pt}
\subsection{Ablation Study}

In this section, we perform ablation experiments to discern the individual contributions of each stream within \TFAname. The results, presented in Tab.~\ref{tab:ablation}, illuminate that both \Intrname and \Intename can autonomously bolster performance by enhancing filters and exploring implicit structures. When \Intrname and \Intename are combined in \TFAname, the performance gains are compounded, showcasing the synergistic benefits of their integration.

\begin{table}[]

\caption{Ablation study on the CIFAR-10 and CIFAR-100.}
\scalebox{0.8}{
\begin{tabular}{cccccccc}
\shline
         \multicolumn{2}{c}{}   & \multicolumn{3}{c}{CIFAR-10}                                                      & \multicolumn{3}{c}{CIFAR-100}                                                     \\ \cline{3-5} \cline{6-8}
     \Intrname & \Intename     & 10                        & 50     & 100                               & 10                        & 50                  & 100               \\ \hline
\rowcolor{lightgray}

\ding{55} & \ding{55}    & 90.83              & 88.38           & 87.97              & 65.74          & 59.48              & 53.24              \\ 
\ding{51} & \ding{55}        & +0.57     & +0.22      & +0.20        & +0.53         & +0.32          & +0.47      \\
\rowcolor{lightgray}
\ding{55} & \ding{51}        & +0.27      & +0.48        & +0.26         & +0.65          & +0.37     & +0.56         \\
\ding{51} & \ding{51}        & +1.74        & +0.91       & +0.67        & +0.99        &  +0.69       & +1.60      \\ \shline
\end{tabular}}
\vspace{-5pt}
\label{tab:ablation}
\end{table}

\begin{figure}[]
  \centering
  \begin{subfigure}{0.495\linewidth}
    \includegraphics[width=1\linewidth]{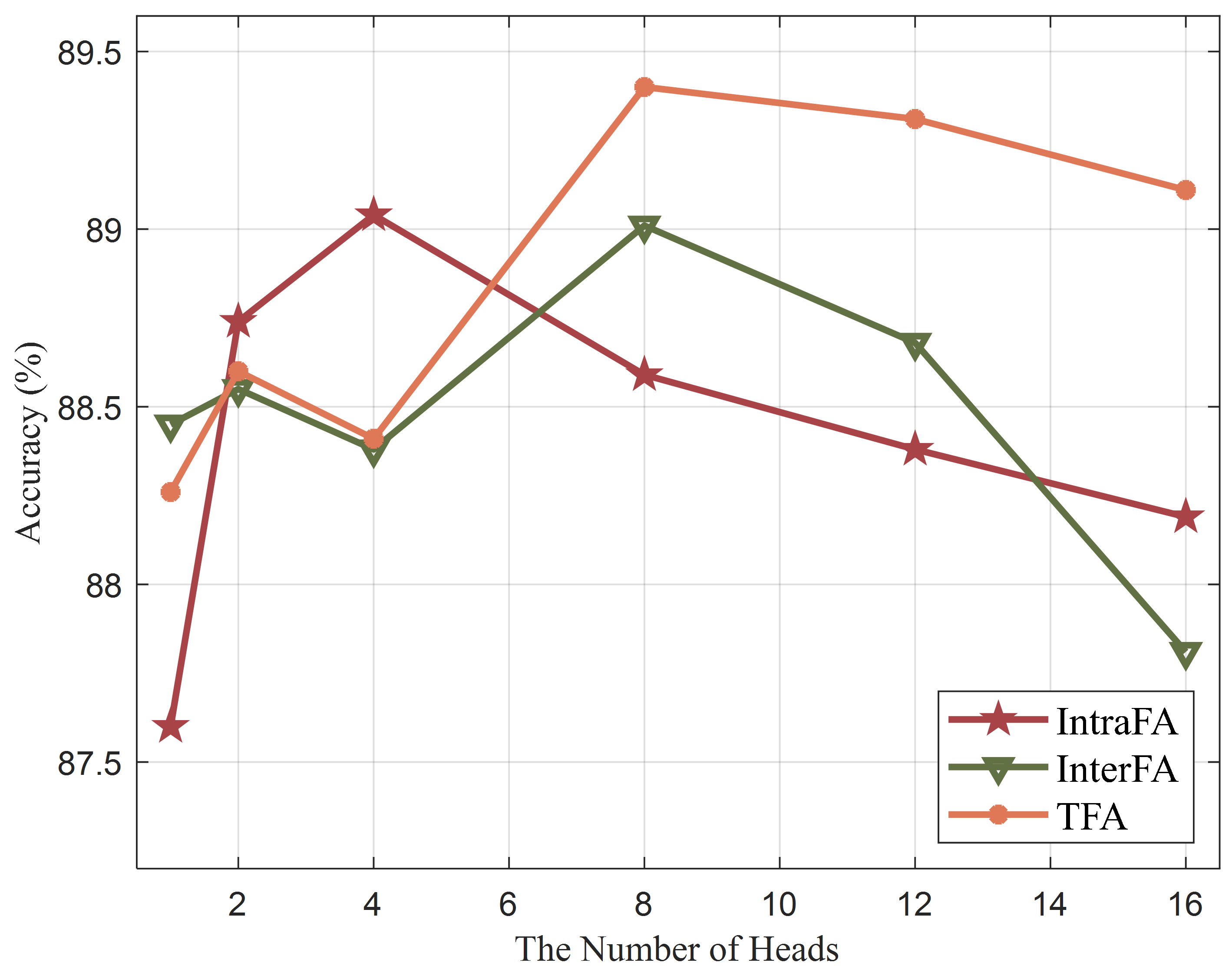}
    \caption{}
    \vspace{-5pt}
  \end{subfigure} 
  \begin{subfigure}{0.495\linewidth}
    \includegraphics[width=1\linewidth]{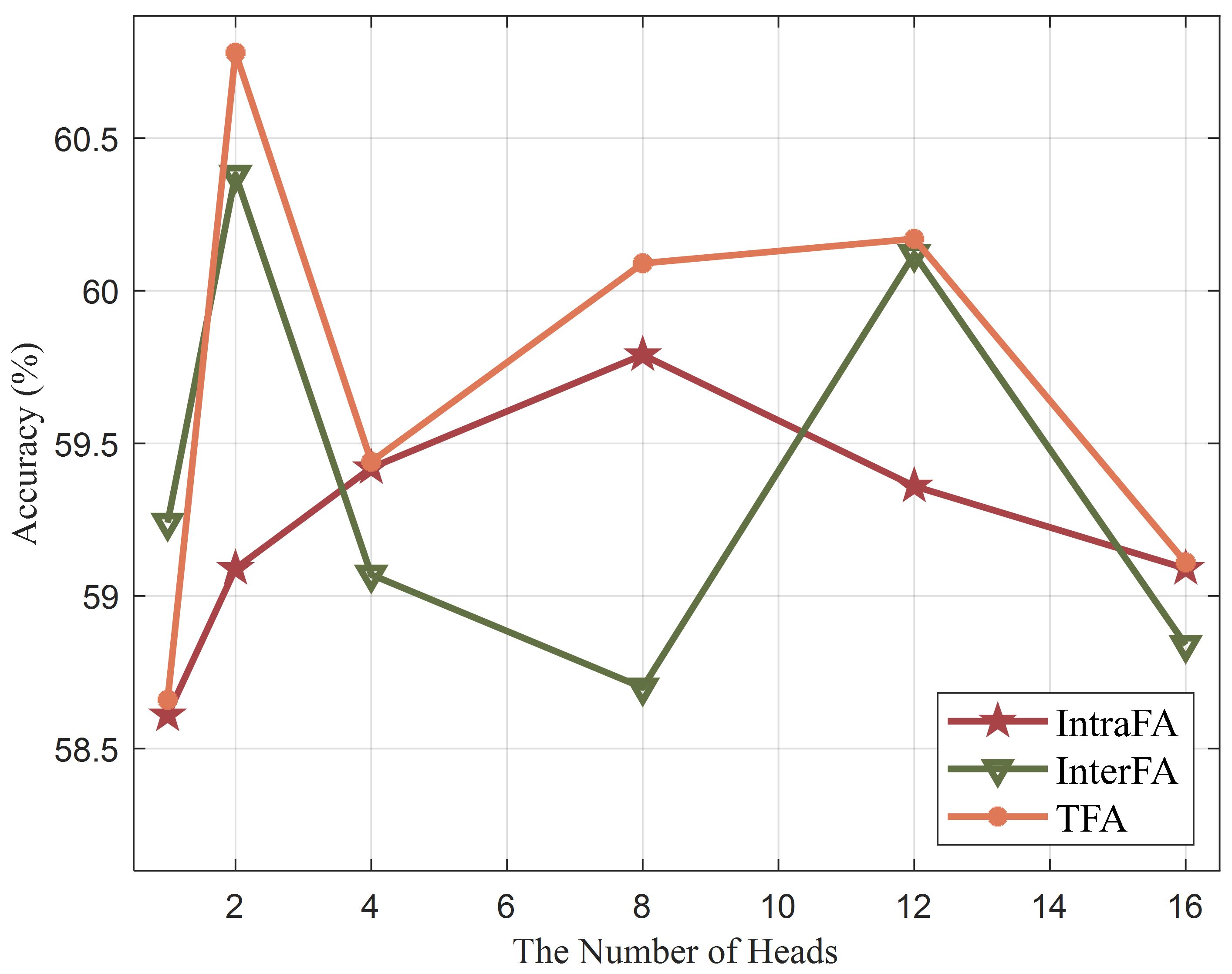}
    \caption{}
    \vspace{-5pt}
  \end{subfigure}
  \caption{The relationship between accuracy and the number of heads. (a)-(b) represent the results on CIFAR-10 and CIFAR-100, respectively.}
  \label{fig:Abal}
  \vspace{-10pt}
\end{figure}

We also investigate the influence of the number of attention heads on performance across 50 clients, and the results are shown in Fig.~\ref{fig:Abal}. Notably, increasing the number of attention heads enhances the model's capacity to consider diverse representations jointly. However, excessive heads make training more challenging and diminish the model's robustness when evaluated on test data.

It's interesting to note that the optimal headcount is similar for both datasets in the case of \Intrname. In contrast, the optimal selection exhibits considerable variation for \Intename and \TFAname. This discrepancy may be attributed to the relatively simplistic structure of client-side models, which might struggle to capture the intricacies of complex tasks. Since \Intename and \TFAname are designed to probe implicit structures, they appear more sensitive to the initially defined client-side architectures. Based on the findings, we opt for a configuration with 2/8 heads for \Intrname and 8/2 heads for \Intename and \TFAname on CIFAR-10/CIFAR-100, respectively.

\begin{figure}
    \centering
    \includegraphics[width=1\linewidth]{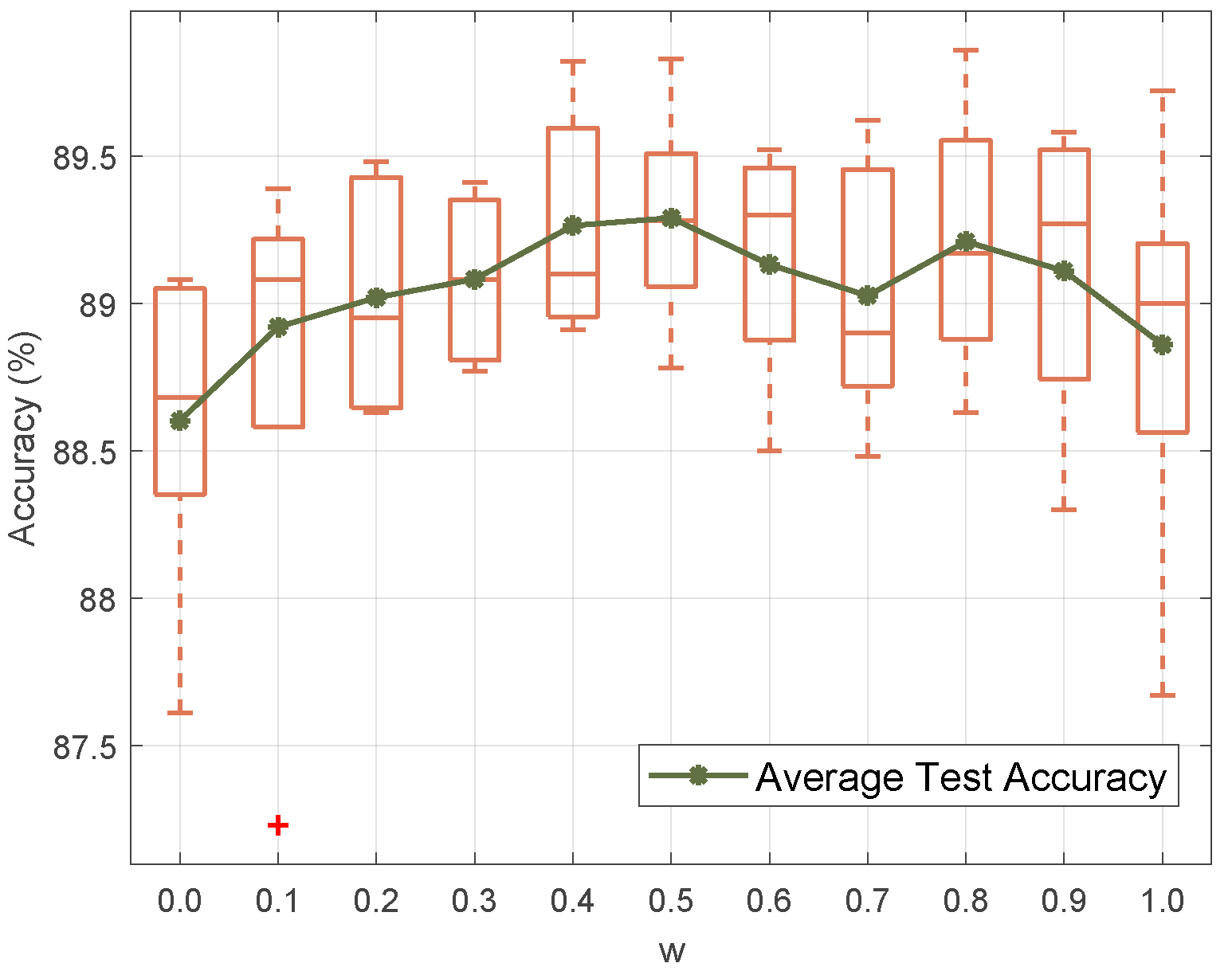}
    \vspace{-5pt}
  \caption{The average test accuracy and boxplot of five experiments about $w$ in  Eq.~\eqref{equation.TFA}.}
  \label{fig:Combination}
  \vspace{-15pt}
\end{figure}

\begin{figure}[]
  \centering
  \begin{subfigure}{0.495\linewidth}
    \includegraphics[width=1\linewidth]{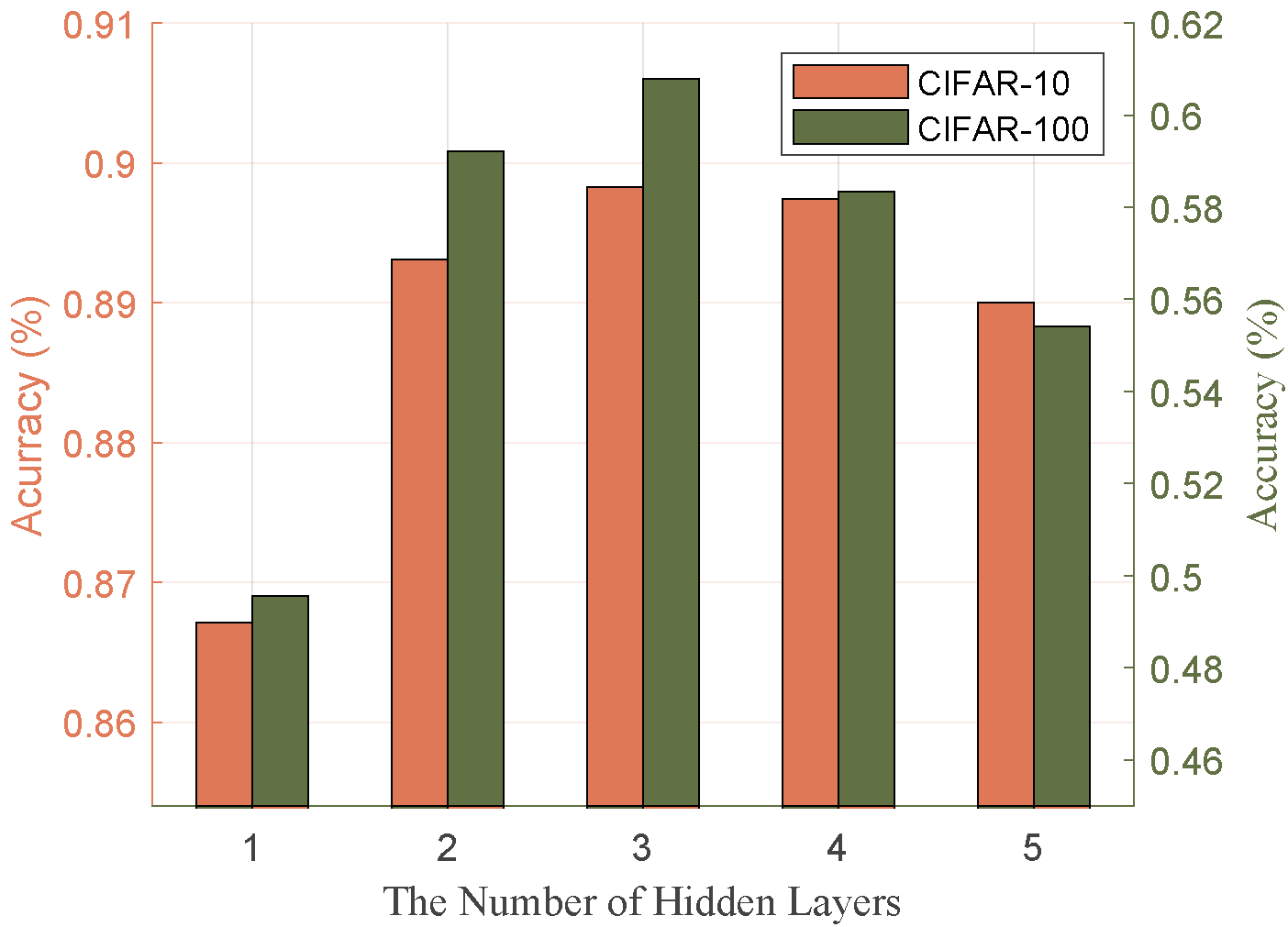}
    \caption{}
    \vspace{-5pt}
    \label{fig:Struc}
  \end{subfigure}
  \begin{subfigure}{0.495\linewidth}
    \includegraphics[width=1\linewidth]{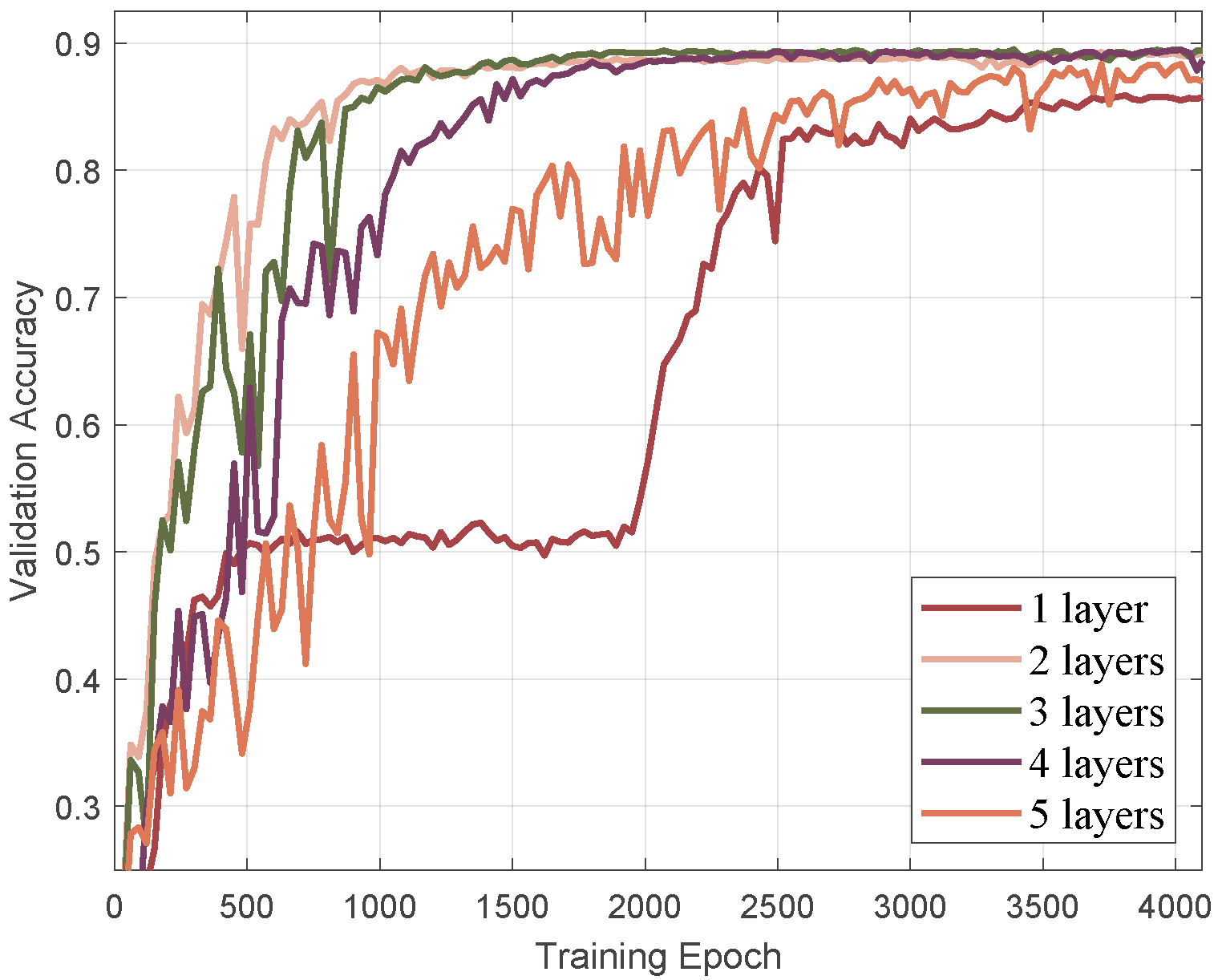}
    \caption{}
    \vspace{-5pt}
    \label{fig:Train}
  \end{subfigure} 
  \caption{The impact of the hypernetwork structure. (a)-(b) represent the test accuracy on CIFAR-10/CIFAR-100 and convergence experiments on CIFAR-10, respectively.}
  \vspace{-15pt}
  \label{fig:Abal_layer}
\end{figure}

In addition, we examine the impact of the weight parameter, denoted as $w$ in Eq.~\eqref{equation.TFA}, with results presented in Fig.~\ref{fig:Combination} across 50 clients using the CIFAR-10 dataset. The findings reveal that combining both \Intrname and \Intename leads to superior performance compared to their utilization ($w=0.0/1.0$). This observation underscores the complementarity of these two modules, with \TFAname capitalizing on their respective strengths for enhanced performance. We propose an empirical setting of $w=0.5$ based on our results as an optimal choice.

Lastly, we explore the impact of hypernetwork structures through experiments, as depicted in Fig.~\ref{fig:Abal_layer}. Specifically, Fig.~\ref{fig:Struc} illustrates related accuracies, while Fig.~\ref{fig:Train} showcases the convergence rates. Notably, a single-layer structure exhibits slower convergence, possibly due to limited representational capacity. Nevertheless, irrespective of the hypernetwork structure, our learning process consistently converges, as the theoretical proof substantiates.

\vspace{-5pt}

\section{Conclusions}

This paper focuses on enhancing filters by redefining the attention mechanism to recalibrate parameters and tailor client-side model structures to specific data. We introduce \methodname for this purpose, emphasizing filter enhancement. \TFAname, a key element within \methodname, is dedicated to refining filters and exploring implicit structures. Furthermore, we propose a strategy to enforce filter orthogonality, diversifying the filter spectrum. An attention-guided pruning approach is presented for customizing local structures and optimizing communication efficiency. Theoretical evidence supports the approximation of filter-aware attention to feature-aware attention, ensuring convergence preservation. Our method distinguishes itself by enhancing performance and reducing communication costs without additional client-side expenses, making filter-aware attention promising for hypernetwork-based FL methods. However, we acknowledge the existing computational constraints that hinder modeling relationships between all filters within intricate models. This remains an open challenge and a focal point for our future work.

{
    \small
    \bibliographystyle{ieeenat_fullname}
    \bibliography{main}
}

\clearpage
\setcounter{page}{1}
\setcounter{table}{0}   
\setcounter{figure}{0}
\setcounter{section}{0}
\setcounter{equation}{0}
\renewcommand{\thetable}{S\arabic{table}}
\renewcommand{\thefigure}{S\arabic{figure}}
\renewcommand{\thesection}{S\arabic{section}}
\renewcommand{\theequation}{S\arabic{equation}}

\maketitlesupplementary
\section{Pseudo Code}
\label{sec:pseudo}

To enhance readers' intuitive grasp of the proposed \methodname, we articulate the primary steps of \methodname in Alg.~\ref{alg:Framwork} using pseudo-code style. For clarity and ease of comprehension, we establish key variables. The training process involves $n$ clients, denoted by $R$ and $K$, for training rounds and local training iterations. $\mathcal{L}_{orth}$ and $\mathcal{L}_{task}$ represent orthogonal regularization and task loss, respectively. The hypernetwork, client-side network, and attention module are parameterized by $\phi$, $\theta_i$, and $\varphi$. Additionally, $\theta'$ signifies enhanced parameters through \TFAname from $\theta$. The proposed AGPS, denoted as $\operatorname{AGPS}(\cdot,p)$, generates a mask matrix $\mat{M}$ to mask $p$\% of unimportant neurons, and $\odot$ signifies the element-wise dot product..

\methodname operates on the server without imposing additional computational burdens on the client. Concurrently, our AGPS masks unimportant neurons, effectively mitigating transmission overhead. Consequently, \methodname can be considered as a plug-and-play module to enhance performance and attain communication efficiency for hypernetwork-based methods. Notably, our approach upholds privacy by eliminating the need for local data sharing during the training process.

\section{Experiments}
\label{sec:addexper}


Our approach demonstrates notable efficacy across both CIFAR-10 and CIFAR-100 datasets. To conduct a more thorough validation of the proposed \methodname, we extend our experiments to include the MNIST dataset~\cite{lecun1998mnist}, ensuring a comprehensive assessment of our method. Consistent with the data partition settings outlined in Section 5.1, we maintain uniformity in experimental conditions, and the results can be found in Tab.~\ref{tab:acc_mnist}. It's evident that our method consistently maintains its promising performance across diverse datasets when compared to the state-of-the-art FL methods. Furthermore, this advantage remains robust even with an increase in the number of clients.

For the input of the framework is the client embedding $\vct{v}$ with a size of 100. However, we are interested in assessing the robustness of the proposed method across varying embedding sizes. In this experiment, we treat a vanilla hypernetwork-based FL method~\cite{shamsian2021personalized} as the baseline, and the results can be found in Tab.~\ref{tab:size}. It can be noticed that the proposed \methodname could significantly improve the performance no matter the size of $\vct{v}_i$, and the performance advantage is particularly pronounced when the embedding size is small.                                   

The above experiments have thoroughly validated the superiority of the proposed \methodname in terms of both accuracy and robustness. These performance improvements do not impose any additional client-side computational costs or increase communication overhead. Therefore, \methodname is well-suited for computational resource limited scenarios, such as those Internet of Things environments.

\begin{algorithm}[t]
  \caption{Main steps of \methodname.}  
  \label{alg:Framwork}
    \textbf{Main Function}   \Comment{Server Executes}\\
    \For{round $r=1,2,...,R$}{
    Randomly select client $i\in\lbrace 1,...,n\rbrace$ \\
    Generate embedding $\boldsymbol{v}_{i}$ \\
    $\theta_i\gets h(\boldsymbol{v}_{i};\phi)$ \\
    $\theta_i' \gets g(\theta_i; \varphi)$ \\
    $\mathcal{L}_{\operatorname{orth}}$ \& Backprop \Comment{Equation~(8)} \\
    $\mat{M}_i\gets \operatorname{AGPS}(\theta^i,p)$  \\
    $\theta_i^M \gets \theta_i'\odot \mat{M}_i$\\ 
    $\Delta \theta_i \gets$\textbf{Local Training}($\theta_i^M$) \\
    Update $\boldsymbol{v}_{i}, \phi,\varphi$ based on $\Delta \theta_i$
    }
\textbf{Local Training$(\theta_i^M)$}   \Comment{$i$-th Client Executes} \\
    $\theta_i^0 \gets \theta_i^M$ \\
    \For{iteration $k=1,2,...,K$}{
        Sample data $x$ and related label $\vct{y}$ from $D_i$\\
        $y'\gets f(\mat{x},\theta)$ \\
        $\mathcal{L}_{\operatorname{task}}(\vct{y}',\vct{y})$ \\
        Backprop \& $\theta_{i}^{k} \leftarrow \theta_{i}^{k-1}-\frac{\partial \mathcal{L}_{\operatorname{task}}}{\partial \theta_{i}^{k-1}}$
    }
    $\Delta \theta_i \gets \theta_i^k - \theta_i^0$ \\
    return $\Delta \theta_i$
    \end{algorithm}
    
\begin{table}[t]
\centering
\vspace{-10pt}
\caption{Test accuracy over 50, 100, and 300 clients on MNIST.}
\vspace{-5pt}
\setlength{\tabcolsep}{12pt}{
\begin{tabular}{rccc}
\shline
          & 50                        & 100                        & 300                                         \\ \hline 
      \rowcolor{lightgray}

FedAvg~\cite{mcmahan2017communication}    & 91.69     &    92.50          &   92.59          \\
FedProx~\cite{li2020federated}   & 92.34    &       92.86          &     92.86           \\
\rowcolor{lightgray}
MOON~\cite{li2021model} & 93.55 & 93.40 & 93.27 \\
FedNova~\cite{wang2020tackling} & 95.84 & 95.51 & 94.71   \\
\rowcolor{lightgray}
FedBN~\cite{li2020fedbn} & 93.15   &   92.33    &   92.21         \\
FedPer~\cite{arivazhagan2019federated}    &    98.08      &    97.37      &  96.08    \\
\rowcolor{lightgray}
FedROD~\cite{chen2021bridging} & 97.32 & 97.97  &  95.74
\\

FedProto~\cite{tan2022fedproto} & 97.17  & 97.55 & 97.68
\\
\rowcolor{lightgray}
pFedHN~\cite{shamsian2021personalized} & 96.79  & 97.20 & 96.14
\\
pFedLA~\cite{ma2022layer} & 97.01 & 96.76 & 95.07
\\ \hline 
\rowcolor{lightgray}
\methodname       & \textbf{98.75}        & \textbf{98.63}         & \textbf{98.17}         \\ \shline
\end{tabular}
}
\vspace{-5pt}
\label{tab:acc_mnist}
\end{table}

\begin{table}[t]
\centering
\vspace{-5pt}
\caption{The performance with different embedding sizes.}
\vspace{-5pt}
\small
\setlength{\tabcolsep}{4pt}{
\begin{tabular}{rcccc}
\shline
        & 50   & 200    &300 & 500         \\ \hline 
      \rowcolor{lightgray}
baseline    & 87.15$\pm$1.36     &    88.57$\pm$1.18  & 88.71$\pm$0.97        &   88.52$\pm$0.98        \\
\methodname       & \textbf{89.84$\pm$0.21}  & \textbf{90.49$\pm$0.29} & \textbf{90.57$\pm$0.13}     & \textbf{90.06$\pm$0.14}\\ \shline
\end{tabular}
}
\vspace{-10pt}
\label{tab:size}
\end{table}


\end{document}